\documentclass[sigconf]{acmart}
\renewcommand\footnotetextcopyrightpermission[1]{}
\usepackage{multirow}
\usepackage{makecell}
\usepackage{lipsum}
\usepackage{algorithm}
\usepackage{algpseudocode}
\usepackage{xcolor}
\usepackage[most]{tcolorbox}

\newcommand{\blockA}[1]{{\color{orange!85!black}#1}}
\newcommand{\blockB}[1]{{\color{blue!70!black}#1}}
\newtcolorbox{graybox}[1][]{
    breakable,
    colback=gray!15,
    colframe=black,
    #1
}

\begin{document}

\title{GeoBlock: Inferring Block Granularity from Dependency Geometry in Diffusion Language Models}

\author{Lipeng Wan}
\affiliation{
  \institution{Xi'an Jiaotong University}
  \city{Xi'an}
  \state{Shaanxi}
  \country{China}
}

\author{Junjie Ma}
\affiliation{
  \institution{Xi'an Jiaotong University}
  \city{Xi'an}
  \state{Shaanxi}
  \country{China}
}

\author{Jianhui Gu}
\affiliation{
  \institution{Xi'an Jiaotong University}
  \city{Xi'an}
  \state{Shaanxi}
  \country{China}
}

\author{Zeyang Liu}
\affiliation{
  \institution{Xi'an Jiaotong University}
  \city{Xi'an}
  \state{Shaanxi}
  \country{China}
}

\author{Xuyang Lu}
\affiliation{
  \institution{Xi'an Jiaotong University}
  \city{Xi'an}
  \state{Shaanxi}
  \country{China}
}

\author{Xuguang Lan}
\affiliation{
  \institution{Xi'an Jiaotong University}
  \city{Xi'an}
  \state{Shaanxi}
  \country{China}
}
\email{xglan@xjtu.edu.cn}

\renewcommand{\shortauthors}{Wan et al.}

\begin{abstract}
Block diffusion enables efficient parallel refinement in diffusion language models, but its decoding behavior depends critically on block size.
Existing block-sizing strategies rely on fixed rules or heuristic signals and do not account for the dependency geometry that determines which tokens can be safely refined together.
This motivates a geometry view of diffusion decoding: \emph{regions with strong causal ordering require sequential updates, whereas semantically cohesive regions admit parallel refinement.}
We introduce GeoBlock, a geometry-aware block inference framework that determines block granularity directly from attention-derived dependency geometry. Instead of relying on predefined schedules or local confidence heuristics, GeoBlock analyzes cross-token dependency patterns to identify geometrically stable refinement regions and dynamically determines appropriate block boundaries during decoding.
By adapting block granularity to the dependency geometry, GeoBlock preserves the parallel efficiency of block diffusion while enforcing dependency-consistent refinement that exhibits autoregressive reliability.
GeoBlock requires no additional training and integrates seamlessly into existing block diffusion architectures.
Extensive experiments across multiple benchmarks show that GeoBlock reliably identifies geometry-consistent block boundaries and improves the accuracy of block diffusion with only a small additional computational budget.
\end{abstract}

\maketitle

\section{Introduction}
\begin{figure}[t]
\begin{center}
\includegraphics[scale=0.55]{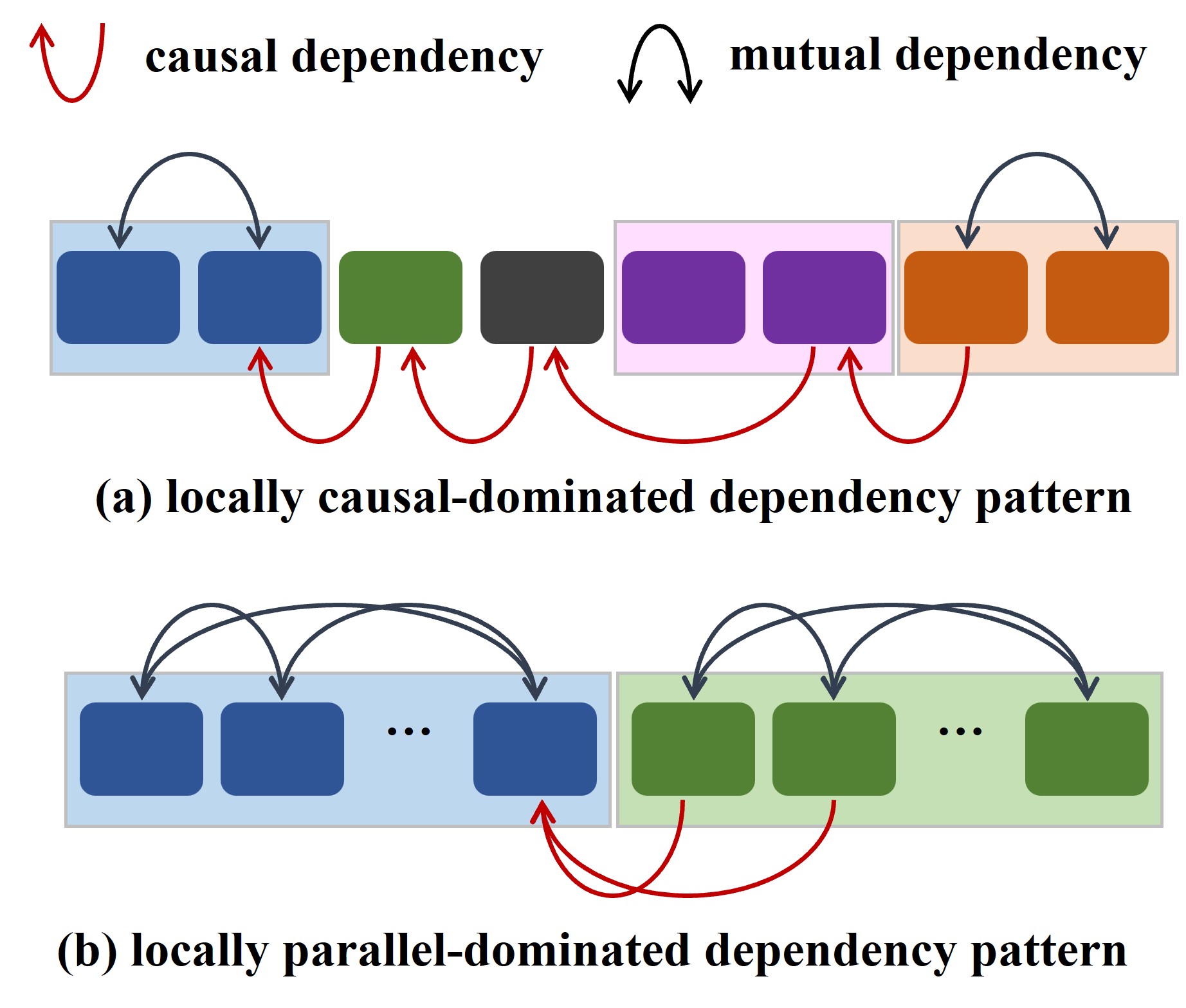}
\caption{Local dependency patterns in natural language.
Tokens with strong mutual interactions are highlighted using the same color.
(a) A locally causal-dominated dependency pattern, where tokens follow a strong directional chain and intermediate tokens exhibit weak local interactions that do not form large, densely connected regions.
Only small-size parallel updates are stable in such regions.
(b) A locally parallel-dominated pattern, where tokens form dense mutual connections, enabling multi-token parallel refinement.
These contrasting structures motivate adaptive block-size selection in dependency-aware diffusion decoding.
\label{intro}}
\end{center}
\end{figure}

Diffusion language models have emerged as an appealing alternative to autoregressive generation, offering parallel and efficient decoding \citep{li2022diffusionlmimprovescontrollabletext, nie2025largelanguagediffusionmodels, lou2024discretediffusionmodelingestimating}.
Block diffusion variants further extend this capability by enabling coarse-grained parallel refinement while retaining compatibility with autoregressive initialization and standard training recipes \citep{sun2025blockwisesftdiffusionlanguage, austin2023structureddenoisingdiffusionmodels}.
These properties have made block diffusion a practical decoding paradigm in recent diffusion-based language models.
However, block diffusion requires the block length to be specified during decoding, and this choice has a substantial impact on decoding behavior.
Small blocks restrict parallelism and slow down convergence, whereas large blocks risk updating many unstable tokens simultaneously, often leading to premature or inconsistent refinement \citep{vonrutte2025scalingbehaviordiscretediffusion, peng2025efficientdiffusionlanguagemodels}.

Although block length is often treated as a decoding hyperparameter, its essential role is to determine how much of the sequence can be updated in parallel at each refinement step \citep{geiping2025efficientparallelsamplersrecurrentdepth, zhong2026parallelismgenerationordermasked}.
This capacity is governed by the underlying dependency structure of the text.
Regions with tight logical progression require sequential conditioning, since each token depends critically on the stability of preceding ones.
In contrast, semantically cohesive regions exhibit dense mutual interactions and can be refined jointly without violating dependency constraints \citep{svete2025reasoningabilitiesmaskeddiffusion, piskorz2025masksdistractingcontextcomprehension}.
Block size should therefore reflect local dependency geometry rather than surface cues such as expected span length.
Length-based heuristics invert the causal structure of language generation: content determines admissible update granularity, not the reverse.
This contrast between locally causal-dominated and locally parallel-dominated dependency patterns is illustrated in Figure~\ref{intro}.

Existing approaches adjust block size using fixed schedules or heuristic signals such as token confidence, denoising volatility, or auxiliary guidance \citep{benhamu2025acceleratedsamplingmaskeddiffusion, israel2025acceleratingdiffusionllmsadaptive, hong2025wideinnarrowoutrevokabledecoding, chen2025confidenceadaptivecoherentdecoding}.
While effective in some cases, these signals characterize token-level uncertainty rather than the relational structure among tokens.
This led us to reconsider what block selection should fundamentally depend on: not individual token confidence, but whether a region forms a self-contained dependency unit under the current decoding state.
A region with high confidence may still contain strong unresolved dependencies that render wide updates unstable, whereas a low-confidence region may already be structurally self-contained and safe to update jointly.
This perspective reveals a structural mismatch between how block size is typically chosen and what the underlying dependency structure of the text actually requires \citep{mohamed2025fastdecodingdiffusionlanguagemodels}.

To bridge this gap, we introduce GeoBlock, a training-free decoding framework that infers block granularity directly from attention-induced dependency geometry during decoding.
GeoBlock formulates block-size selection as a local boundary inference problem driven by attention-induced dependency geometry.
Self-attention provides a model-consistent proxy for dependency structure, allowing the decoder to determine whether a candidate region can be updated jointly without violating dependency constraints \citep{li2025diffusionlanguagemodelsknow, kong2025acceleratingdiffusionllminference, mo2025decodinglargelanguagediffusion}.
GeoBlock assigns each candidate region a closure score that balances internal coupling and past anchoring against dependency leakage to the unresolved future, and selects the rightmost split whose score lies within a tolerance $\delta$ of the maximum.
This rule yields the largest geometry-consistent block under the current dependency structure: strong future leakage favors smaller blocks, while internally self-contained regions permit larger parallel updates.
As a result, update granularity adapts continuously to local dependency geometry while remaining fully compatible with standard block diffusion decoding. \citep{sun2025blockwisesftdiffusionlanguage, xu2025lopascalingdllminference}.

This work makes three contributions.
\begin{itemize}
    \item First, we introduce a structural-geometric perspective on block diffusion decoding, viewing block granularity as a consequence of dependency geometry rather than a predefined schedule or heuristic signal.
    \item Second, we propose GeoBlock, a training-free, geometry-aware block boundary inference method that determines adaptive block granularity directly from attention-induced dependency structure during decoding.
    \item Third, experiments across multiple benchmarks validate dependency geometry as an effective principle for block selection: GeoBlock consistently improves decoding performance while incurring only a modest increase in computation (approximately 11\% additional Number of Function Evaluations  (NFE) ).
\end{itemize}

\section{Background}
\subsection{Masked Diffusion Language Models}
Diffusion-based language models adapt the corruption--reconstruction paradigm to discrete sequences by replacing tokens with an absorbing masking state and learning to invert this transformation \citep{sohldickstein2015deepunsupervisedlearningusing, austin2023structureddenoisingdiffusionmodels, lou2024discretediffusionmodelingestimating}. 
In masked diffusion language models (MDLMs), the forward process progressively maps tokens to a special \texttt{[MASK]} symbol according to a predefined schedule, while the reverse process iteratively predicts the clean sequence using a bidirectional Transformer conditioned on partially observed context \citep{sahoo2024simpleeffectivemaskeddiffusion, nie2025largelanguagediffusionmodels}. 
At each refinement step, the model maintains a partially observed sequence and predicts, in parallel, a categorical distribution over all masked positions conditioned on the full visible context. 
An updated intermediate sequence is obtained by sampling or selecting token predictions, followed by a remasking operation that determines which positions remain unresolved for further refinement.

This formulation induces a globally coupled decoding process in which all token updates are performed in parallel and conditioned on the current full-sequence state. 
Unlike autoregressive generation, decoding is not restricted to a fixed token order: at each step, the model refines an arbitrary subset of positions while leveraging bidirectional context.
As a result, diffusion decoding proceeds through iterative global refinement rather than through a strictly sequential chain of token emissions, enabling parallel updates and flexible information flow across the sequence.

Given an original sequence $x_{0}$ and its corrupted version $x_{t}$ obtained by masking at ratio $t$, MDLMs optimize a masked reconstruction loss that predicts only the tokens hidden by the corruption process:
\begin{equation}
\mathcal{L}(\theta)
=
-\,\mathbb{E}_{t,\,x_{0},\,x_{t}}
\left[
\frac{1}{t}
\sum_{\ell=1}^{L}
\mathbf{1}\!\left[x_{t}^{(\ell)} \in \mathcal{M}\right]
\log p_{\theta}\!\left(x_{0}^{(\ell)} \mid x_{t}\right)
\right],
\end{equation}
where $\mathcal{M}$ denotes the masking state. 
This objective trains the model to denoise masked positions using visible context and can be interpreted as a discrete diffusion evidence lower bound \citep{ou2025absorbingdiscretediffusionsecretly, shi2025simplifiedgeneralizedmaskeddiffusion}.

Although MDLMs support efficient multi-token refinement, standard decoding treats all masked positions uniformly and does not adapt to the heterogeneous dependency structure of natural language. 
Some regions follow tight causal or logical ordering and benefit from sequential stabilization, while semantically cohesive spans can be refined jointly without introducing instability. 
Standard masked diffusion provides no mechanism to distinguish between these regimes, motivating structured update strategies that adapt refinement granularity to dependency geometry, such as variable block sizes in block diffusion decoding \citep{tian2025nexttokennextblockprincipledadaptation, sun2025blockwisesftdiffusionlanguage}.

\begin{figure*}[t]
\begin{center}
\includegraphics[width=\textwidth]{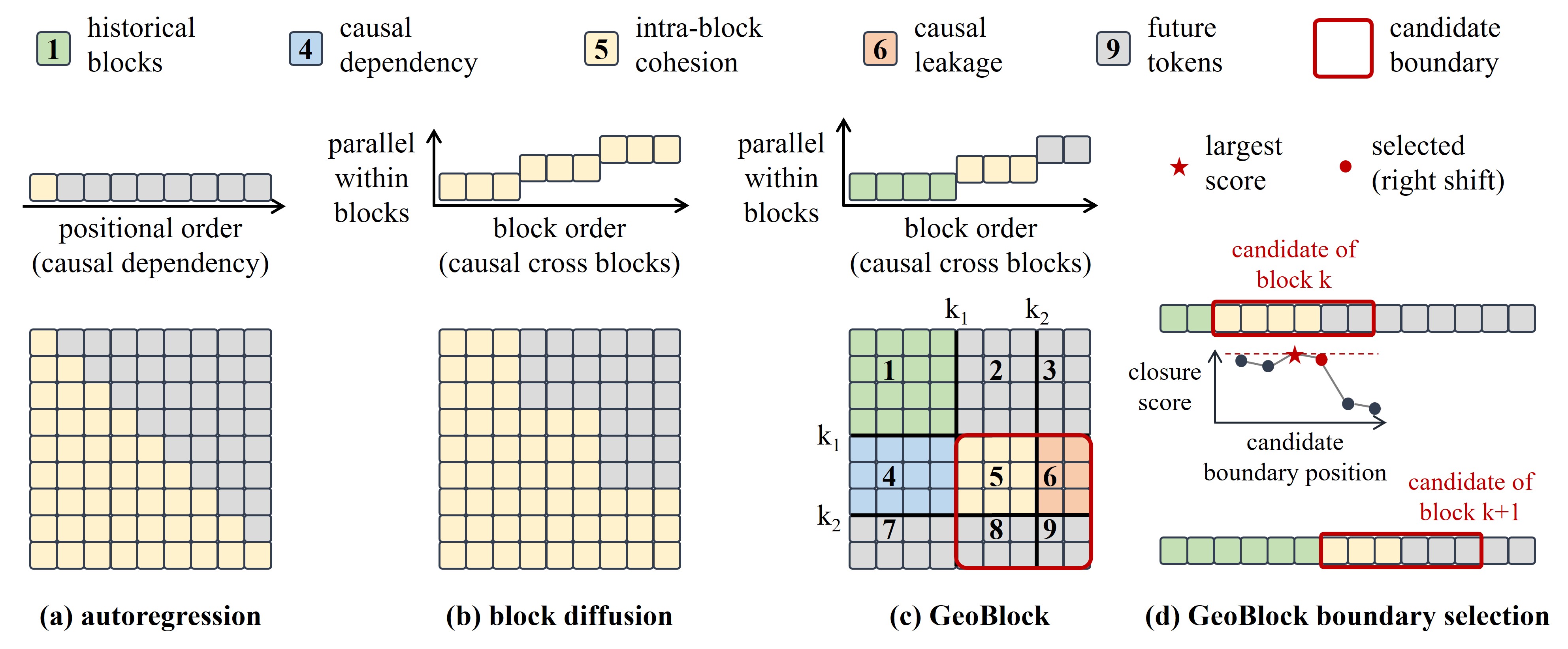}
\caption{
Dependency geometry of decoding under different regimes.
(a) Autoregressive decoding induces a strictly sequential dependency structure, yielding a triangular attention pattern.
(b) Block diffusion groups tokens into contiguous blocks, allowing bidirectional interactions within blocks while preserving causal ordering across blocks.
(c) GeoBlock decomposes frontier attention into structured regions over historical tokens, candidate blocks, and future tokens, highlighting internal coupling, past conditioning, and future leakage.
(d) Geo boundary selection evaluates candidate boundaries using a closure score and selects the right-shifted boundary within a tolerance of the maximum.
\label{fig:method}}
\end{center}
\end{figure*}

\subsection{Blockwise Strategies for Diffusion Decoding}

Masked diffusion language models refine all masked positions in parallel,
implicitly assuming that tokens can be updated independently at each denoising step.
In practice, natural language exhibits structured dependencies that are unevenly distributed across tokens.
Some regions follow strict causal or syntactic ordering, where token stability depends
strongly on preceding context, while others form semantically cohesive spans whose
internal dependencies can be resolved jointly.
Blockwise diffusion methods \citep{arriola2025blockdiffusion} introduce structural
control by restricting each denoising step to a contiguous update region.
This design enables bidirectional conditioning within the block, preserves causal
ordering across blocks, and establishes a continuum between fully parallel diffusion
and strictly sequential decoding.

Subsequent work explores how the update granularity should be selected or adapted
during inference.
Existing strategies adjust block size using learned policies \citep{ctrldiff2025}
or training-free heuristics derived from confidence, entropy, or volatility signals
\citep{adablock2025, benhamu2025acceleratedsamplingmaskeddiffusion}.
Other approaches vary the effective update region implicitly, such as committing
high-confidence prefixes \citep{sdlm2025} or expanding a dynamic active region along
semantic boundaries \citep{wavefront2025}.
These methods demonstrate that update granularity has a direct impact on refinement
stability, convergence behavior, and generation quality.
However, their selection criteria rely on uncertainty proxies or auxiliary objectives,
rather than directly characterizing the dependency relations among tokens being
updated.

Beyond fixed-canvas decoding, variable-length diffusion and insertion-based denoising methods \citep{daedal2024, flexmdm2024, editdiffusion2024} allow the active token set to expand or contract during generation.
These approaches underscore a common insight: effective diffusion decoding requires dynamically determining which tokens participate in each refinement step, rather than adhering to a fixed update pattern.
However, they regulate participation through sequence-length dynamics, rather than through inference over dependency structure within a fixed canvas.

Taken together, prior work highlights the importance of adaptive control over update regions in diffusion-based decoding.
However, existing approaches determine update granularity using heuristic uncertainty signals, learned decision policies, or sequence-length operations, rather than directly inferring whether a candidate set of tokens forms a coherent dependency unit under the current decoding state.
This motivates treating block selection as a structural inference problem, deriving update regions from observable dependency structure rather than from surface-level uncertainty alone.

A more detailed survey of related methods is provided in
Appendix~\ref{app:related_work}.

\section{Method}
\subsection{Problem Setup: Dependency Geometry in Block Diffusion}
Let $x_{1:L}$ denote a length-$L$ token sequence. 
Autoregressive decoding induces a strictly sequential dependency structure in which each token $x_i$ is generated conditioned on the prefix $x_{<i}$, yielding a one-dimensional causal dependency graph and a triangular attention pattern with $A_{ij}=0$ for $j>i$.
While this structure preserves maximal causal fidelity, it enforces a fully sequential update regime and precludes parallel refinement.

Block diffusion lifts this restriction by shifting the unit of conditioning from individual tokens to contiguous blocks.
The sequence is partitioned into blocks $\{B_k\}$ processed in causal order, while allowing unrestricted bidirectional attention within each block.
Formally, for tokens $i,j\in B_k$, attention weights $A_{ij}$ are unconstrained, whereas cross-block attention obeys a block-level causal order.
As illustrated in Figure~\ref{fig:method}, this induces a two-scale dependency geometry: causal dependencies across blocks and parallel dependency resolution within blocks.
Block size thus determines the maximal set of tokens whose mutual dependencies are resolved simultaneously at each refinement step. Figure~\ref{fig:method} provides a schematic overview of this dependency geometry and the resulting block boundary inference process.

\paragraph{Block size as a structural assumption.}
From this perspective, block size is not merely a decoding hyperparameter governing a speed--quality trade-off.
Rather, it encodes a structural assumption on conditional independence: selecting a block $B$ asserts that tokens within $B$ can be jointly updated without violating essential dependency constraints.
When $|B|=1$, the decoding process reduces to a strictly sequential chain; when $|B|$ is large, the decoder assumes that tokens in $B$ form a mutually consistent dependency unit.
This assumption may be invalid if internal dependencies within the block have not yet stabilized, leading to unreliable joint updates.

Existing block diffusion methods typically determine $|B|$ using external heuristics such as fixed schedules, confidence scores, or entropy-based criteria.
However, these signals primarily reflect token-level uncertainty rather than the relational dependency structure among tokens.
As a result, they do not directly address the central structural question posed by block diffusion decoding: whether a candidate region forms a self-contained dependency unit under the current decoding state.
We therefore reformulate block selection as a dependency inference problem grounded in attention-derived observations of the model's decoding geometry.

\paragraph{Frontier-based dependency decomposition.}
We consider decoding at a frontier position $y$, where tokens $x_{1:y}$ have been partially refined and tokens $x_{>y}$ remain uncommitted.
For any candidate split position $x<y$, this frontier induces a partition of indices into three contiguous sets:
\[
H = \{1,\dots,x\}, \quad
C = \{x+1,\dots,y\}, \quad
F = \{y+1,\dots\}.
\]

Let $A\in\mathbb{R}^{L\times L}$ denote a self-attention matrix produced by the model at the current decoding step, serving as an observable proxy for the underlying dependency structure.
Let $A^{(y)} = A_{[1:y,\;1:L]}$ denote its restriction to queries within the current frontier.
Under the partition $(H,C,F)$, the matrix $A^{(y)}$ decomposes into nine submatrices $A_{U,V}$ with $U,V\in\{H,C,F\}$, corresponding to the regions illustrated in Figure~\ref{fig:method}.
Each submatrix captures a distinct class of dependencies between query tokens in $U$ and key tokens in $V$.
While these patterns may vary across heads or layers, their geometric roles in determining dependency closure remain consistent.

Not all regions are informative for assessing whether the candidate set $C$ can be treated as a valid block.
Regions involving only history or future tokens (e.g., $A_{H,H}$ or $A_{F,F}$) reflect background structure independent of the current block decision, and regions with queries in $H$ primarily encode how previously committed tokens attend to later positions.
In contrast, the regions $A_{C,C}$, $A_{C,H}$, and $A_{C,F}$ directly characterize the dependency geometry of the candidate set.
Here, $A_{C,C}$ captures internal coupling within $C$, $A_{C,H}$ reflects conditioning on the past, and $A_{C,F}$ measures dependency leakage toward future, unresolved tokens.
A candidate region that forms a self-contained dependency unit should therefore exhibit strong internal coupling and past anchoring, while maintaining minimal dependence on the future.
This geometric criterion forms the basis of our closure score, formalized in the next section.

\begin{algorithm}[t]
\caption{GeoBlock boundary inference with fused attention}
\label{alg:geoblock_boundary}
\begin{algorithmic}[1]
\Require
Frontier window $[s,e)$ of length $L$;
selected layers $\mathcal{L}$ with weights $\{w_\ell\}$;
top-$k$ salient heads per layer;
minimum block length $m$;
right-shift tolerance $\delta$.
\Ensure Selected boundary $x^\star$.

\vspace{2pt}
\Statex \textbf{1. Multi-layer salient attention fusion}
\State $A \gets 0$
\For{$\ell \in \mathcal{L}$}
    \State extract ROI attentions $\{A^{(\ell,h)}\}_{h=1}^H$
    \State select salient heads $\mathcal{H}_\ell$ by total attention mass
    \State $\bar A^{(\ell)} \gets \mathrm{mean}_{h\in\mathcal{H}_\ell} A^{(\ell,h)}$
    \State $A \gets A + w_\ell \cdot \bar A^{(\ell)}$
\EndFor

\vspace{2pt}
\Statex \textbf{2. Dependency-closure scoring}
\For{$x=m,\dots,L-1$}
    \State define $H=\{1,\dots,x-m\}$,\; $C=\{x-m+1,\dots,x\}$,\; $F=\{x+1,\dots,L\}$
    \State $S_{C\to C}\gets \sum_{i\in C}\sum_{j\in C} A_{ij}$ \Comment{internal coupling}
    \State $S_{C\to H}\gets \sum_{i\in C}\sum_{j\in H} A_{ij}$ \Comment{past anchoring}
    \State $S_{C\to F}\gets \sum_{i\in C}\sum_{j\in F} A_{ij}$ \Comment{future leakage}
    \State $\mathrm{Score}(x)\gets
    \dfrac{S_{C\to C}+\alpha S_{C\to H}}
    {S_{C\to C}+\alpha S_{C\to H}+S_{C\to F}}$
\EndFor

\vspace{2pt}
\Statex \textbf{3. Right-shift boundary selection}
\State $S_{\max}\gets \max_x \mathrm{Score}(x)$,\;\;
$\tau\gets S_{\max}-\delta$
\State $x^\star\gets \max\{x:\mathrm{Score}(x)\ge\tau\}$
\State \Return $x^\star$
\end{algorithmic}
\end{algorithm}

\subsection{Dependency-Aware Block Selection}

\paragraph{Dependency observation and closure criterion.}
At a decoding frontier $y$, we consider candidate regions of the form
$C=\{x+1,\dots,y\}$ for split positions $x<y$.
Let $H=\{1,\dots,x\}$ denote the resolved prefix and $F=\{y+1,\dots\}$ the
unresolved suffix.
Attention patterns produced during decoding provide observable evidence of
how tokens within the current frontier interact and condition on future tokens.

Given an attention matrix $A\in\mathbb{R}^{L\times L}$, we quantify dependency
strength between two index sets $U,V$ by the accumulated attention mass
\[
S_{U\to V}=\sum_{i\in U}\sum_{j\in V}A_{ij}.
\]
For a candidate region $C$, three quantities are particularly informative:
$S_{C\to C}$ measuring internal coupling within $C$,
$S_{C\to H}$ measuring conditioning on the past, and
$S_{C\to F}$ measuring dependency leakage toward unresolved future tokens.
A region that forms a self-contained dependency unit should therefore exhibit
strong internal coupling and past anchoring while maintaining minimal
dependence on the future.

We measure the dependency closure of $C$ using the following closure score:
\[
\mathrm{Score}(x)=
\frac{
S_{C\to C}+\alpha\,S_{C\to H}
}{
S_{C\to C}+\alpha\,S_{C\to H}+S_{C\to F}
},
\]
where $\alpha\in[0,1]$ balances internal cohesion and conditioning on the past.
This score favors regions whose dependencies can be resolved jointly under the
current prefix while penalizing residual reliance on unresolved future tokens.

\paragraph{Aggregated dependency inference.}
Attention observations from different layers and heads provide complementary
views of the dependency structure.
Rather than inferring boundaries from individual observations, we aggregate
them into a unified dependency estimate and perform boundary inference on the
fused representation.

Let $\mathcal{L}$ denote a selected set of attention layers.
For each $\ell\in\mathcal{L}$, we compute a layer-level estimate
$\bar A^{(\ell)}$ by averaging its most informative heads within the current
frontier window.
These layer-level observations are combined through a fixed weighted sum,
yielding a fused attention matrix
\[
A=\sum_{\ell\in\mathcal{L}} w_\ell\,\bar A^{(\ell)},\qquad
\sum_{\ell\in\mathcal{L}}w_\ell=1,
\]
which provides a unified observation of dependency geometry around the frontier.
All dependency quantities in Eq.~(3) are computed from this fused attention map,
resulting in a single score profile $\mathrm{Score}(x)$ over candidate splits.

\paragraph{Boundary selection.}
Given the score profile over candidate splits, we infer the block boundary
directly from the aggregated dependency landscape.
Selecting the strict maximizer can yield overly conservative blocks due to
minor score fluctuations.
To enable maximal admissible expansion while preserving structural stability,
we adopt a right-shift rule:
among all candidate splits whose score lies within $\delta$ of the maximum,
we select the rightmost one,
\[
x^\star
=
\max\big\{
x:\;\mathrm{Score}(x)\ge \mathrm{Score}_{\max}-\delta
\big\}.
\]
The resulting region $C=\{x^\star+1,\dots,y\}$ defines the block used for the
current refinement step.
When the inferred block collapses to a single token, the procedure reduces to
the sequential regime as a limiting case.

\begin{figure*}[t]
\begin{center}
\includegraphics[width=\textwidth]{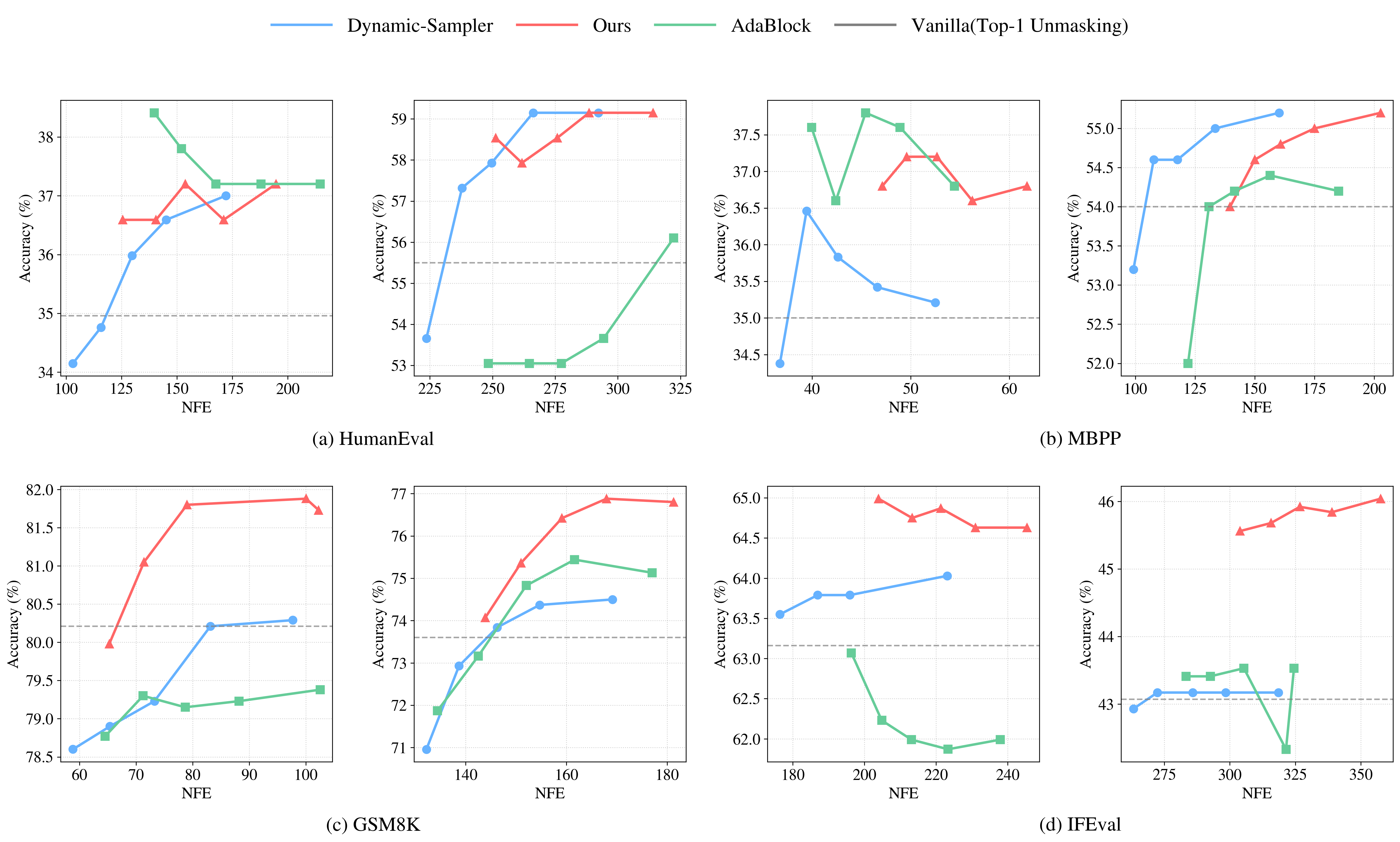}
\caption{Accuracy versus number of function evaluations (NFE) on HumanEval, MBPP, GSM8K, and IFEval under Dream-7B and LLaDA-8B backbones.
\label{exp1}}
\end{center}
\end{figure*}

\begin{table*}[htb]
  \caption{Accuracy (Acc) and decoding cost (NFE) across benchmarks. 
  Bold indicates best accuracy per row. 
  Red/green arrows denote improvement/decline of GeoBlock over Dynamic at matched block size.}
  \label{tab:main}
  \begin{tabular}{lccccccccccccc}
    \toprule
    & & \multicolumn{3}{c}{Vanilla} & \multicolumn{3}{c}{Dynamic} &  \multicolumn{3}{c}{AdaBlock} &  \multicolumn{3}{c}{GeoBlock} \\
    \cmidrule(lr){3-5} \cmidrule(lr){6-8} \cmidrule(lr){9-11}  \cmidrule(lr){12-14}
    Benchmark & & B16 & B32 & B64 & B16 & B32 & B64 &$B_0 16$ &$B_0 32$ & $B_0 64$ & $B_{max}16$ & $B_{max}32$ & $B_{max}64$\\
    \midrule
    \multicolumn{14}{c}{Dream-7B}\\
    \midrule

    \multirow{2}{*}{GSM8K}
    & ACC & 74.60 & 73.62 & 74.83 & 75.13 & 74.37 & 74.68 & 74.98 & 75.44 & 73.31 
          & \textbf{77.79}{\textcolor{red}{$\uparrow$}} 
          & \textbf{76.88}{\textcolor{red}{$\uparrow$}} 
          & \textbf{77.86}{\textcolor{red}{$\uparrow$}} \\
    & NFE & 256 & 256 & 256 & 151.72 & 154.69 & 160.83 & 167.44 & 161.62 & 161.26 & 169.80 & 167.96 & 168.92 \\

    \multirow{2}{*}{HumanEval}
    & ACC & 57.93 & 55.49 & 57.93 & 53.66 & \textbf{59.15} & 55.49 & 51.83 & 53.66 & 57.93 
          & 55.49{\textcolor{red}{$\uparrow$}} 
          & \textbf{59.15} 
          & 55.49 \\
    & NFE & 512 & 512 & 512 & 258.98 & 266.19 & 267.44 & 303.15 & 294.35 & 289.89 & 292.10 & 288.53 & 292.10 \\

    \multirow{2}{*}{MBPP}
    & ACC & 54.20 & 54.00 & \textbf{55.80} & \textbf{55.80} & 55.00 & 55.60 & 54.20 & 54.40 & 55.40 
          & 52.40{\textcolor{green}{$\downarrow$}} 
          & 51.80{\textcolor{green}{$\downarrow$}} 
          & 52.40{\textcolor{green}{$\downarrow$}} \\
    & NFE & 512 & 512 & 512 & 131.83 & 133.39 & 128.75 & 174.86 & 156.43 & 141.22 & 174.17 & 175.00 & 174.17 \\

    \multirow{2}{*}{IFEval}
    & ACC & 42.14 & 43.07 & 39.93 & 45.84 & 43.17 & 42.57 & 44.24 & 42.33 & 40.41 
          & \textbf{46.88}{\textcolor{red}{$\uparrow$}} 
          & 45.84{\textcolor{red}{$\uparrow$}} 
          & \textbf{46.88}{\textcolor{red}{$\uparrow$}} \\
    & NFE & 512 & 512 & 512 & 293.20 & 298.42 & 296.69 & 329.25 & 321.45 & 317.82 & 339.52 & 338.96 & 339.52 \\

    \midrule
    \multicolumn{14}{c}{LLaDA-8B}\\
    \midrule

    \multirow{2}{*}{GSM8K}
    & ACC & \textbf{82.18} & 80.21 & 79.08 & 82.11 & 80.21 & 81.35 & 81.58 & 79.23 & 79.91 
          & 81.27{\textcolor{green}{$\downarrow$}} 
          & \textbf{81.88}{\textcolor{red}{$\uparrow$}} 
          & 79.45{\textcolor{green}{$\downarrow$}} \\
    & NFE & 256 & 256 & 256 & 88.59 & 83.13 & 79.79 & 93.97 & 88.18 & 84.42 & 98.63 & 100.05 & 85.95 \\

    \multirow{2}{*}{HumanEval}
    & ACC & 37.20 & 34.76 & 36.59 & 34.15 & 36.59 & 35.37 & 36.59 & 37.20 & 33.54 
          & \textbf{37.80}{\textcolor{red}{$\uparrow$}} 
          & 36.59 
          & \textbf{37.80}{\textcolor{red}{$\uparrow$}} \\
    & NFE & 512 & 512 & 512 & 163.45 & 145.12 & 130.80 & 203.21 & 187.93 & 169.69 & 200.39 & 171.10 & 149.48 \\

    \multirow{2}{*}{MBPP}
    & ACC & 36.20 & 35.00 & 30.20 & 37.60 & 35.42 & 20.20 & 37.80 & 37.60 & 27.80 
          & \textbf{40.00}{\textcolor{red}{$\uparrow$}} 
          & 36.60{\textcolor{red}{$\uparrow$}} 
          & 34.20{\textcolor{red}{$\uparrow$}} \\
    & NFE & 512 & 512 & 512 & 60.84 & 46.60 & 40.89 & 64.28 & 48.92 & 42.50 & 84.61 & 56.24 & 47.38 \\

    \multirow{2}{*}{IFEval}
    & ACC & 64.88 & 63.22 & 58.78 & 62.29 & 63.96 & 63.67 & \textbf{66.67} & 61.87 & 59.59 
          & \textbf{66.67}{\textcolor{red}{$\uparrow$}} 
          & 64.63{\textcolor{red}{$\uparrow$}} 
          & 64.87{\textcolor{red}{$\uparrow$}} \\
    & NFE & 512 & 512 & 512 & 215.50 & 207.25 & 199.33 & 239.60 & 223.33 & 206.65 & 250.91 & 230.97 & 219.23 \\

    \bottomrule
  \end{tabular}
\end{table*}

\paragraph{Computational efficiency.}
The proposed block inference procedure operates entirely on attention matrices
already produced during decoding and requires no additional forward passes.
Dependency quantities $S_{U\to V}$ are accumulated incrementally within a bounded
frontier window, resulting in linear complexity with respect to the candidate
block size.
Under practical horizons (e.g., $|C|\leq 64$), this overhead remains negligible
relative to the cost of self-attention computation and subsequent block
diffusion updates.
Block inference is performed locally at each decoding frontier and does not
require global sequence segmentation.

\section{Experiments}
\subsection{Experimental Setup}
\paragraph{Backbones.}
We evaluate GeoBlock on representative diffusion language models spanning both scratch-trained and autoregressive-adapted families, including LLaDA-1.5 and Dream-v0-Base-7B \citep{zhu2025llada15variancereducedpreference, ye2025dream7bdiffusionlarge}.
These models follow the masked discrete diffusion formulation introduced in prior work on diffusion language modeling \citep{nie2025largelanguagediffusionmodels, austin2023structureddenoisingdiffusionmodels}.
For instruction-following settings, we additionally report results on instruction-tuned variants when applicable.
All methods reuse the same pretrained backbones without any retraining or finetuning.

\paragraph{Decoding framework.}
All methods are evaluated within the same block-diffusion decoding pipeline.
We keep the underlying denoising model, prompting format, and stopping rules fixed across methods, and vary only the decoding control mechanism.
GeoBlock is implemented as a drop-in replacement for the block-size scheduler, operating within the standard blockwise refinement loop without modifying model parameters or training procedures.
We compare against representative block-adaptive decoding baselines, including AdaBlock, which dynamically adjusts block size using token-level semantic confidence signals to enable adaptive multi-token commitment~\citep{adablock2025}.
All experimental settings follow the default inference configuration of Fast-dLLM to ensure consistent and reproducible decoding behavior across methods~\citep{wu2025fastdllmtrainingfreeaccelerationdiffusion}.
All methods use the same key--value caching strategy during decoding to ensure a fair comparison of computational cost.

\paragraph{Benchmarks and metrics.}
We conduct experiments on standard reasoning, instruction-following, and code-generation benchmarks.
Mathematical reasoning is evaluated on GSM8K and MATH \citep{cobbe2021trainingverifiers, hendrycks2021measuring}, instruction-following on IFEval \citep{zhou2023ifeval}, and code generation on HumanEval and MBPP \citep{chen2021evaluating, austin2021programsynthesis}.
Performance is measured using answer accuracy for math and instruction-following tasks and pass@1 for code generation.
Decoding efficiency is reported using the number of function evaluations (NFE), defined as the total number of model forward passes required to produce a complete output.
Where relevant, we also report wall-clock throughput for completeness.
We provide additional qualitative case studies illustrating the inferred block structures and refinement behavior of GeoBlock in Appendix~\ref{app:case}.

\subsection{Main Results}

\begin{table}[htb]
\caption{Block length and additional NFE ratio of GeoBlock.}
\label{tab:block_overhead}
\centering
\begin{tabular}{l|cc|cc}
\toprule
& \multicolumn{2}{c|}{Dream} & \multicolumn{2}{c}{LLaDA} \\
Benchmark 
& \makecell{Block\\Length}
& \makecell{Extra\\NFE} 
& \makecell{Block\\Length} 
& \makecell{Extra\\NFE} \\
\midrule
GSM8K     & $13.72\pm5.87$    & 0.08   & $13.42 \pm 5.69$  & 0.08 \\
HumanEval & $14.33\pm4.95$ & 0.12  & $14.99\pm 5.26$ & 0.14 \\
MBPP      & $16.54\pm4.79$    & 0.15   & $19.76\pm 5.96$ & 0.29 \\
MATH    & $11.81\pm5.14$    & 0.14   & $11.67\pm 5.31$ & 0.15 \\
IFEval    & $17.86\pm6.05$    & 0.07   & $18.65\pm 6.35$ & 0.08 \\
\bottomrule
\end{tabular}
\end{table}

\paragraph{Comparison with existing block decoding strategies.}
Table~\ref{tab:main} compares GeoBlock with vanilla block decoding, dynamic confidence-based decoding, and AdaBlock under matched decoding budgets. 
Across both Dream-7B and LLaDA-8B, GeoBlock achieves the best or comparable accuracy in most settings while maintaining similar decoding cost, indicating a favorable accuracy--efficiency trade-off. 
In particular, GeoBlock tends to outperform heuristic block-selection strategies at medium block sizes, where reliable multi-token commitment is most critical.

On Dream-7B, GeoBlock delivers the highest accuracy on GSM8K and IFEval across matched block configurations and remains competitive on HumanEval, suggesting improved refinement stability for reasoning and long-form generation. 
Performance on MBPP is slightly below dynamic decoding, indicating that aggressive multi-token commitment is less advantageous for short program synthesis tasks.

On LLaDA-8B, GeoBlock remains robust across benchmarks and block configurations. 
It achieves the highest accuracy on GSM8K at the medium block setting and consistently improves over dynamic decoding on HumanEval and MBPP, while attaining the best performance on IFEval across all block sizes. 
Overall, these results show that incorporating dependency geometry into block construction yields more reliable refinement regions and improved generation quality under comparable or moderately increased decoding cost.

\paragraph{Block length and computational overhead.}
To better understand the computational behavior of GeoBlock, Table~\ref{tab:block_overhead} reports the average inferred block length and the additional decoding cost introduced by dependency-aware boundary construction. 
Across both Dream and LLaDA backbones, GeoBlock produces moderately sized blocks (typically 13–19 tokens), indicating that dependency-aware closure expansion tends to form compact yet meaningful refinement regions rather than overly large segments. 
The additional decoding cost remains modest across benchmarks, with extra NFE ratios mostly within 7–15\% for reasoning and instruction-following tasks, and slightly higher on MBPP due to longer inferred blocks. 
These results suggest that GeoBlock achieves improved accuracy primarily through more reliable multi-token commitment rather than substantially increasing decoding computation, maintaining a favorable efficiency profile across tasks.

\paragraph{Accuracy--NFE trade-off under varying confidence thresholds.}
While Table~\ref{tab:main} reports representative operating points, we further examine the accuracy--efficiency trade-off of GeoBlock under different commitment aggressiveness. 
We vary the confidence threshold from 0.75 to 0.95, producing five operating points that span a wide range of decoding costs. 
Lower thresholds enable earlier commitment and fewer refinement steps, whereas higher thresholds lead to more conservative decoding and increased NFE.

Figure~\ref{exp1} shows the resulting accuracy--NFE curves across HumanEval, MBPP, GSM8K, and IFEval for both Dream-7B and LLaDA-8B. 
Across most tasks and thresholds, GeoBlock traces a Pareto-dominant or comparable frontier relative to Dynamic decoding and AdaBlock. 
On reasoning-heavy benchmarks such as GSM8K and IFEval, GeoBlock consistently achieves higher accuracy at similar or lower NFE, indicating that dependency-aware boundary selection enables more reliable multi-token commitment. 
On HumanEval and MBPP, GeoBlock remains competitive across the full threshold range and often attains higher peak accuracy as the decoding budget increases.

Overall, the results show that GeoBlock maintains a favorable and stable accuracy--efficiency trade-off across a broad range of decoding budgets rather than at a single tuned operating point.

\subsection{Ablation Studies}

To better understand the design choices of GeoBlock, we conduct ablation studies on the key components that govern dependency-aware block inference. 
Specifically, we analyze (i) the anchoring coefficient $\alpha$ in the closure score, which balances internal cohesion and past conditioning; 
(ii) the right-shift tolerance $\delta$, which controls how aggressively blocks expand; 
and (iii) the selection and weighting of attention layers used for dependency estimation. 
All experiments are conducted under matched decoding budgets to ensure fair comparisons between configurations and to isolate the effect of each design component on the accuracy--efficiency trade-off.

\paragraph{Effect of anchoring coefficient.}
Table~\ref{tab:alpha_ablation} studies the impact of the anchoring coefficient $\alpha$.
Setting $\alpha=0$ removes the contribution of past anchoring and reduces the closure score to a purely internal–future trade-off, while larger values increase reliance on historical conditioning.
Across both LLaDA-8B and Dream-7B, incorporating past anchoring ($\alpha>0$) consistently improves decoding accuracy compared to $\alpha=0$, indicating that stable conditioning on resolved tokens is important for identifying reliable refinement regions and preventing premature commitment.
Intermediate values ($\alpha=0.25$ or $0.5$) yield the best accuracy--efficiency trade-off across models, achieving strong accuracy with moderate decoding cost.
Excessively large anchoring ($\alpha=1.0$) slightly degrades performance, suggesting that overemphasizing historical conditioning can overly constrain block growth and reduce opportunities for parallel refinement.
Overall, these results indicate that balancing structural cohesion with moderate anchoring is critical for stable and efficient block inference.

\paragraph{Effect of right-shift tolerance.}
Table~\ref{tab:delta_ablation} examines the right-shift tolerance $\delta$, which determines how close a candidate boundary must be to the optimal closure score before being selected.
Increasing $\delta$ leads to progressively larger inferred block lengths, enabling more aggressive parallel refinement and reducing NFE by committing larger token groups at each step.
However, overly large tolerance introduces mild accuracy degradation due to premature block expansion and reduced refinement flexibility.
We observe that moderate values ($\delta=0.1$) provide the best overall trade-off, achieving the highest accuracy while maintaining low decoding cost and stable block sizes.
This confirms that allowing limited right-shift expansion stabilizes block inference by avoiding overly conservative boundary selection while preserving dependency consistency across refinement steps.

\paragraph{Effect of layer and weight configurations.}
Table~\ref{tab:layer_weight_gsm8k} evaluates different attention layer selections and fusion weights for dependency estimation.
We compare early-layer, mid-layer, and high-layer combinations with multiple weighting schemes.
Results show that mid-to-high semantic layers (e.g., 16-21-26) consistently provide the strongest performance, suggesting that dependency geometry useful for block inference emerges more clearly in later layers where semantic relationships are more stable.
Balanced weight assignments across selected layers yield robust performance, while heavily skewed weighting offers no consistent advantage.
These findings indicate that GeoBlock is not overly sensitive to precise layer or weight choices as long as dependency estimation is drawn from semantically mature layers, supporting the robustness of the proposed geometry-based block inference mechanism.

\begin{table}[htb]
  \caption{Effect of the anchoring coefficient under matched decoding budgets.}
  \label{tab:alpha_ablation}
  \centering
  \begin{tabular}{l|l|cccc}
    \toprule
    Benchmark & Metric & 0 & 0.25 & 0.5 & 1.0 \\
    \midrule
    \multirow{2}{*}{Dream-7B}
        & Acc &    76.65   &   \textbf{76.88}    &    \textbf{76.88}   &    76.73   \\
        & NFE &   169.40  &   \textbf{164.15}    &   167.96    &   172.57   \\
        \midrule
    \multirow{2}{*}{LLaDA-8B}
        & Acc & 81.19 & 81.35 & \textbf{81.88} & 80.82 \\
        & NFE &  \textbf{98.72}   &  102.51  &   100.05   &  103.05  \\
    \bottomrule
  \end{tabular}
\end{table}

\begin{table}[htb]
  \caption{Effect of the right-shift tolerance on boundary selection.}
  \label{tab:delta_ablation}
  \centering
  \begin{tabular}{l|ccc}
    \toprule
    $\delta$ & Acc & NFE & Block Len. \\
    \midrule
    0    & 80.74 & 115.03 & $6.59\pm 3.12$ \\
    0.05 & 80.74 & 102.99  & $10.28 \pm 4.21$ \\
    0.1  & \textbf{81.88} & 100.05  &  $13.42\pm 5.69$ \\
    0.2  & 80.21 & \textbf{98.64}  & $19.57\pm 9.33$ \\
    \bottomrule
  \end{tabular}
\end{table}

\begin{table}[htbp]
\centering
\caption{Layer and weight configurations on GSM8K (LLaDA, threshold=0.9, $\delta=0$).}
\label{tab:layer_weight_gsm8k}
\begin{tabular}{ccccc}
\toprule
\textbf{Layer} & \textbf{Weight} & \textbf{Acc} & \textbf{NFE} \\
\midrule
\multirow{3}{*}{2-6-10}
& 0.1, 0.3, 0.6      & 79.15 & 123.33 \\
& 0.6, 0.3, 0.1      & 81.12 & 127.56 \\
& 0.333, 0.333, 0.334 & 80.67 & 126.26 \\
\midrule
\multirow{3}{*}{16-21-26}
& 0.1, 0.3, 0.6      & 80.59 & 100.61 \\
& 0.6, 0.3, 0.1      & 81.12 & 105.35 \\
& 0.333, 0.333, 0.334 & \textbf{81.88} & \textbf{100.05} \\
\midrule
\multirow{3}{*}{20-24-26}
& 0.1, 0.3, 0.6     & 80.74 & 103.03  \\
& 0.6, 0.3, 0.1      & 79.98 & 103.71 \\
& 0.333, 0.333, 0.334 & 80.97 & 101.20 \\
\bottomrule
\end{tabular}
\end{table}

\subsection{Dependency and Block Boundary Visualization}
\begin{figure}[ht]
\begin{center}
\includegraphics[width=\columnwidth]{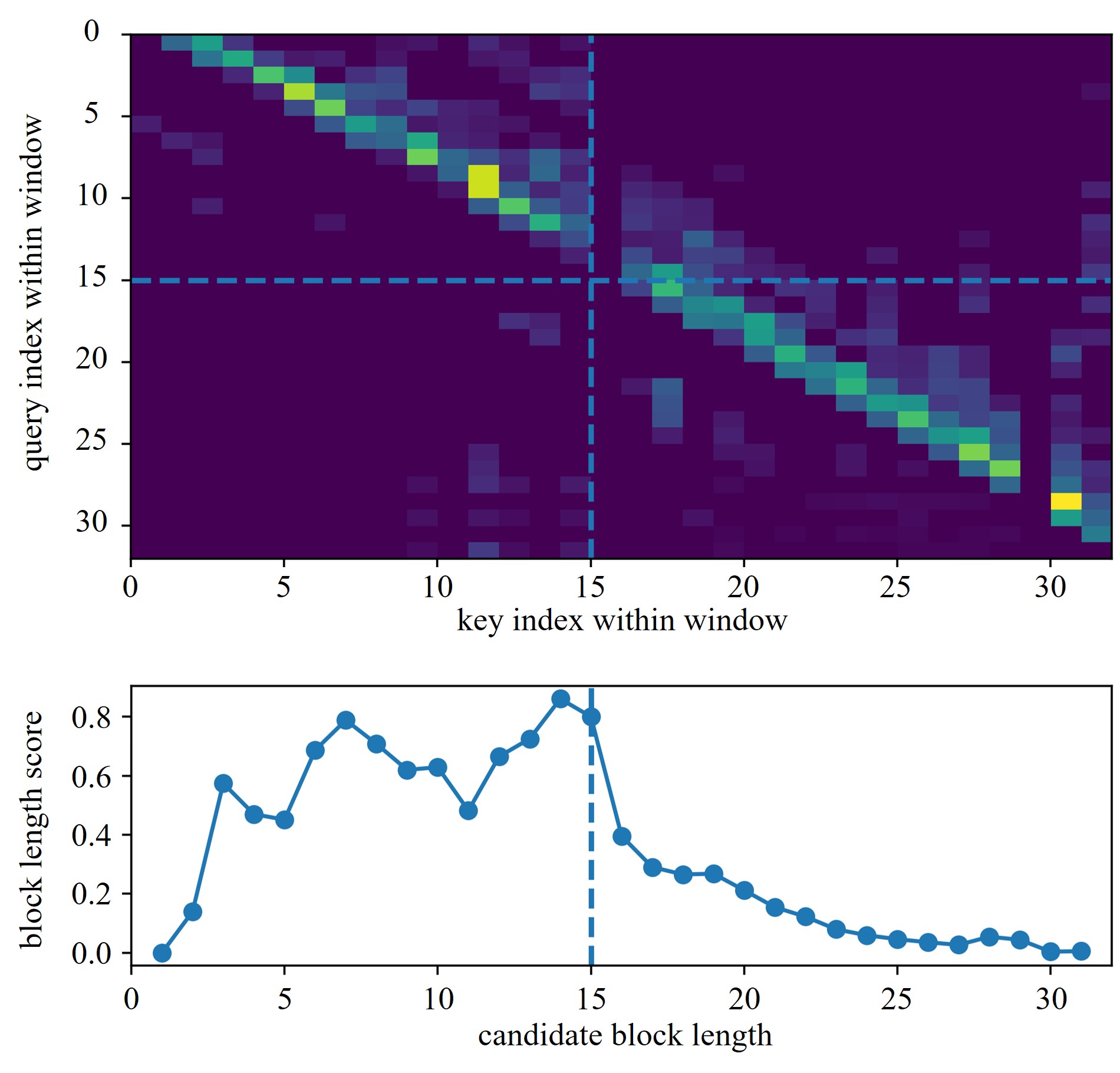}
\caption{
Block boundary inference under dependency geometry during diffusion decoding.
Top: Fused attention map within the frontier window, revealing internal cohesion and cross-boundary dependencies.
Bottom: Block closure score as a function of candidate boundary position.
The selected boundary (dashed line) corresponds to the largest candidate block whose closure score remains near-optimal, allowing stable commitment of a structurally coherent token subset.
}\label{fig:vis_block}
\end{center}
\end{figure}

To better understand how dependency-aware block inference operates during decoding, 
we visualize the attention structure and corresponding block closure scores 
at representative decoding frontiers.
Since GeoBlock determines block boundaries from attention-derived dependency geometry, 
this analysis provides a direct view of how inferred blocks align with the evolving structural dependencies of the sequence.

At selected refinement steps, we extract the fused attention map within the current frontier window 
and compute the closure score for each candidate boundary as defined in Section~3.2.
Figure~\ref{fig:vis_block} shows both the attention pattern and the resulting score profile,
with the selected block boundary indicated by a dashed line.

The visualization reveals a consistent correspondence between attention-derived dependency structure 
and inferred block boundaries.
When tokens within a region exhibit strong mutual interactions and limited dependence on unresolved future tokens, 
the closure score remains high across an extended span, allowing the inferred block to expand.
In contrast, when substantial dependency leakage toward future tokens is present, 
the score drops sharply and the inferred block contracts accordingly.
These observations confirm that GeoBlock adapts update granularity in accordance with the model’s internal dependency geometry,
enabling stable commitment in sequentially constrained regions while preserving parallel refinement where structure permits.

\section{Conclusion}

We presented GeoBlock, a geometry-aware block inference framework for diffusion language model decoding that determines block granularity directly from attention-derived dependency structure. 
Instead of relying on fixed schedules or token-level confidence heuristics, GeoBlock formulates block selection as a structural inference problem and identifies geometry-consistent refinement regions through a dependency-closure criterion. 
By adapting block granularity to the evolving dependency geometry of the sequence, the method enables parallel refinement where structural independence permits while preserving stability in sequentially constrained regions. 
GeoBlock is training-free, introduces only modest computational overhead, and integrates seamlessly into existing block diffusion decoding pipelines.

Experiments across reasoning, code generation, and instruction-following benchmarks show that GeoBlock consistently achieves strong or improved generation quality while maintaining a favorable accuracy--efficiency trade-off across backbones and block configurations. 
Overall, this work highlights the importance of structural and relational signals in diffusion decoding and suggests that modeling dependency geometry offers a principled path toward more adaptive and reliable parallel generation in diffusion language models.


\bibliographystyle{ACM-Reference-Format}
\bibliography{sample-base}

\appendix

\section{Related Works}\label{app:related_work}
\subsection{Adaptive Block Size in Block Diffusion Decoding}
Block diffusion enables parallel refinement by updating contiguous token blocks in a causal order, with block size determining the granularity of joint updates.
Early formulations adopt a fixed block size.
Block Diffusion \citep{arriola2025blockdiffusion} performs diffusion within blocks of predetermined length and predicts blocks autoregressively, exposing a fundamental trade-off between parallelism and refinement stability but leaving block granularity static.

Recent work explores adaptive block-size selection during inference.
AdaBlock-dLLM \citep{adablock2025} adjusts block size using training-free heuristics derived from token-level confidence and volatility signals.
CtrlDiff \citep{ctrldiff2025} instead treats block-size selection as a learned decision problem, training a policy network to predict block length under a quality--efficiency objective.
These approaches allow block size to vary over time, but rely on uncertainty proxies or learned rewards rather than explicit modeling of inter-token dependency structure \citep{wang2025remaskingdiscretediffusionmodels, kim2025klassklguidedfastinference}.

Other methods induce variable block boundaries implicitly.
Sequential Diffusion Language Models \citep{sdlm2025} generate variable-length subsequences and commit the longest high-confidence prefix at each step, effectively yielding dynamic block boundaries through prefix selection.
From a more principled perspective, From Next-Token to Next-Block \citep{tian2025nexttokennextblockprincipledadaptation} frames block size as a continuous generalization of autoregressive decoding, interpreting the AR regime as the limiting case where block size equals one.
This view highlights block granularity as a structural choice rather than a purely computational parameter \citep{luxembourg2025planspeeddilatedscheduling}.

Several studies further analyze how training and architectural factors influence block-size behavior.
Blockwise SFT \citep{sun2025blockwisesftdiffusionlanguage} shows that attention masking and future leakage during training affect which block sizes remain stable at inference.
At scale, LLaDA2.0 \citep{bie2025llada20scalingdiffusionlanguage} adopts staged or progressive block sizes, providing empirical evidence that block granularity can evolve across decoding stages or model capacities \citep{bie2025llada20scalingdiffusionlanguage}.

In contrast to these approaches, which determine block size via heuristic uncertainty measures, learned policies, or implicit prefix rules, our method infers block boundaries directly from the dependency structure among tokens.
By leveraging attention-induced dependency geometry, we select block sizes based on whether a candidate region forms a self-contained dependency unit under the current decoding state, rather than on confidence or volatility alone.
This enables adaptive block selection that is structurally grounded and fully training-free \citep{ni2026flexibilitytraparbitraryorder}.

\subsection{Variable-Length and Dynamic-Canvas Diffusion Models}
Another line of research addresses the fixed-length limitation of diffusion language models by allowing the token canvas to change during generation.
Rather than assuming a predefined sequence length, these methods enable diffusion models to determine how many tokens should be generated or edited as part of the denoising process \citep{chen2025dlmonediffusionlanguagemodels, xue2025anyordergptmaskeddiffusion}.
These formulations relax the fixed-length assumption of standard masked diffusion and enable more flexible generation regimes.

Several approaches focus on training-free or inference-time length adaptation.
DAEDAL \citep{daedal2024} dynamically extends the sequence before or during denoising based on completion signals, enabling diffusion models to infer missing content length.
Seed Diffusion \citep{seeddiffusion2024} reformulates the corruption process in terms of expected insertions and deletions, allowing sequence length to evolve naturally with the noise schedule.
Related work also explores dynamic refinement schedules and length-aware inference mechanisms that adjust generation trajectories online \citep{miao2025contextawareinitializationreducinggenerative, shen2026improvingthroughputdiffusionbasedlarge}.

A related family of methods defines diffusion over edit operations.
Edit-based diffusion models \citep{editdiffusion2024} treat insertion and deletion as primitive stochastic edits, with noise levels corresponding to expected edit distance.
Flexible masked diffusion models, such as FlexMDM \citep{flexmdm2024}, alternate between inserting masked tokens and predicting their content, supporting dynamic canvases during denoising.
Other infilling approaches \citep{flexinfilling2024} reconstruct masked spans with variable length, often using span-level couplings or transport-based objectives.
Together, these approaches treat generation as a sequence editing process rather than fixed-position denoising.
Recent diffusion formulations further explore edit-based or transport-inspired generation processes that naturally accommodate evolving sequence structure during denoising \citep{zhou2025coevolutionarycontinuousdiscretediffusion, peng2025pathplanningmaskeddiffusion}.

These methods address the question of \emph{how many tokens should exist or be edited} during diffusion-based generation.
In contrast, blockwise decoding assumes a fixed canvas and focuses on \emph{how many existing tokens should be updated jointly at each refinement step}.
Our work operates on this latter axis, using dependency structure to adapt block granularity without modifying sequence length or the diffusion process itself \citep{liu2025wedlmreconcilingdiffusionlanguage}.
This distinction places our method orthogonal to dynamic-length diffusion approaches and highlights block granularity as a structural property of decoding rather than a function of sequence length.

\subsection{Attention-Based Structural Signals for Decoding}

A growing body of work studies attention patterns in language models as indicators of underlying structural or relational properties.
In diffusion language models, recent analyses show that attention is not merely a byproduct of denoising, but exhibits systematic geometric structure.
Attention Sinks in Diffusion Language Models \citep{attentionsinks2025} identify dynamically evolving sink tokens during diffusion, indicating that dependency patterns shift across refinement steps.
Sparse attention studies such as SparseD \citep{sparsed2025} further reveal substantial heterogeneity across attention heads, with distinct heads consistently capturing different structural roles.
These findings suggest that attention patterns in diffusion models encode meaningful dependency information rather than noise \citep{jin2025rolediscretenessdiffusionllms, piskorz2025masksdistractingcontextcomprehension, song2025sparsedllmacceleratingdiffusionllms}.

Blockwise diffusion architectures explicitly impose geometric structure on attention through masking.
Block Diffusion \citep{arriola2025blockdiffusion} defines a block-causal attention pattern with unrestricted intra-block interactions, shaping how dependencies are resolved during decoding.
Subsequent analysis in Blockwise SFT \citep{sun2025blockwisesftdiffusionlanguage} shows that attention leakage across block boundaries significantly affects inference behavior, highlighting the importance of forward and backward dependency signals in blockwise settings.
Together, these works indicate that attention geometry plays a central role in governing which tokens can be stably refined together and how structural constraints propagate during denoising \citep{zhou2025coevolutionarycontinuousdiscretediffusion, nguyentri2025attentionneedkvcache, ma2025dkvcachecachediffusionlanguage}.

Beyond diffusion models, attention has also been explored as a decoding-time control signal in autoregressive generation.
Prior studies use attention concentration or entropy to guide adaptive decoding decisions, such as early stopping or branching \citep{shen2025staticcutoffsoneshotdynamic, fu2025bitsroundsparalleldecoding}, and structural probing literature demonstrates that individual attention heads often align with syntactic or dependency relations.
While these approaches do not operate in diffusion decoding regimes, they provide further evidence that attention encodes relational structure exploitable at inference time, as also suggested by recent diffusion decoding analyses \citep{lee2025lookaheadunmaskingelicitsaccurate, mo2025decodinglargelanguagediffusion}.

In contrast to prior work, which primarily analyzes attention patterns or uses them as auxiliary control signals, our method leverages attention-induced dependency geometry to directly infer block granularity during diffusion decoding.
Rather than modifying attention mechanisms or decoding objectives, we treat attention observations as structural evidence for determining whether a candidate region forms a self-contained dependency unit, enabling dependency-aware block adaptation without additional training.

\section{Implementation Details} 

\paragraph{Decoding configuration.}
Across all benchmarks, we fix the maximum generation length to $L=512$ tokens and use the standard block-diffusion inference procedure.
All methods share identical sampling temperatures, prompt templates, and termination conditions to ensure fair comparison.
When a fixed block size is required, we evaluate multiple choices $B\in\{16,32,64\}$.
In contrast, GeoBlock infers an effective block size dynamically at each refinement step while remaining within the same decoding loop.

\paragraph{Baselines.}
We compare against a vanilla block-diffusion baseline and dynamic decoding variants that adapt update behavior using token-level uncertainty signals.
Cache-accelerated variants are also included where KV caching is enabled.
All baselines are implemented under the same decoding framework and evaluated with identical budgets and hardware settings.

\section{Case Studies}\label{app:case}
\begin{graybox}[title={Case 1: LLaDA-8B, GSM8K, Generation Length 256}]

\textbf{Question}

Janet’s ducks lay 16 eggs per day. She eats three for breakfast every morning and bakes muffins for her friends every day with four. She sells the remainder at the farmers' market daily for \$2 per fresh duck egg. How much in dollars does she make every day at the farmers' market?

\medskip

\textbf{Ground-truth Answer}

Janet sells 16 - 3 - 4 = <<16-3-4=9>>9 duck eggs a day.
She makes 9 * 2 = <<9*2=18>>18 every day at the farmer’s market.

\noindent\rule{\linewidth}{0.5pt}

\textbf{GeoBlock, max block length = 32, $\delta=0.1$}

\blockB{To determine how much Janet makes every day at the farmers' market, we need to follow} \blockA{ these steps:}

\blockA{1. Calculate the total number of eggs laid by Janet's ducks per day.}

\blockB{2. Subtract the number of eggs Janet eats and bakes for her friends from the} \blockA{ total number of eggs laid.}

\blockA{3. Determine the number of eggs she sells each day.}

\blockB{4. Calculate the revenue from selling the eggs} \blockA{.}

\blockA{Let's start with the first step:}

\blockB{Janet's ducks lay 16 eggs per day.}

\blockB{Next, we subtract} \blockA{ the number of eggs she eats and bakes for her friends:}

\blockA{- She} \blockB{ eats 3 eggs for breakfast.}

\blockB{- She bakes 4 eggs for her friends.} \blockA{So, the number of eggs she sells each day is:}
\[
\blockB{16 - 3 - 4 = 9}
\]
\blockB{Now, we calculate the} \blockA{ revenue from selling the eggs:}\blockA{Janet sells each egg for \$2. Therefore,} \blockB{ the daily revenue from selling 9 eggs is:}
\[
\blockB{9 \times 2 = }\blockA{18}
\]
\blockA{Thus} \blockB{, the amount of money Janet makes every day at the farmers' market is }\boxed{\blockA{18}} \blockA{ dollars.}
\\

\textbf{Block length}: $16.0 \pm 3.9$. \textbf{Number of blocks}: 16.

\noindent\rule{\linewidth}{0.5pt}

\textbf{GeoBlock, max block length = 64, $\delta=0.05$}

\blockB{To determine how much Janet makes every day at the farmers' market, we need to follow these steps:}

\blockB{1. Calculate the total number of eggs laid by Janet's ducks per} \blockA{day.}

\blockA{2. Subtract the number of eggs Janet eats and bakes for her friends from the total number of eggs laid.}

\blockA{3. Calculate the revenue} \blockB{from selling the remaining eggs at the farmers' market.}

\blockB{Step 1: Calculate the total number of eggs laid}  \blockA{by Janet's ducks per day.}

\blockA{Janet's ducks lay 16 eggs per day.}

\blockA{Step 2: Subtract the number of eggs Janet eats and bakes for} \blockB{her friends from the total number of eggs laid.}

\blockB{Janet eats 3 eggs for breakfast and bakes 4 eggs for her friends. So, the number of eggs left to sell is}
\blockA{:}
\[
\blockA{16 - 3 - 4 = 9}
\]

\blockA{Step 3: Calculate the revenue from selling the remaining eggs at the farmers' market.}

\blockA{Janet sells each egg for \$2} \blockB{. Therefore, the daily revenue is:}
\[
\blockB{9 \times 2} \blockA{ = 18}
\]
\blockB{So, Janet makes } \boxed{\blockB{18}} \blockB{ in dollars every day} \blockA{.}
\\

\textbf{Block length}: $25.6 \pm 13.81$. \textbf{Number of blocks}: 10.

\noindent\rule{\linewidth}{0.5pt}

\textbf{GeoBlock, max block length = 64, $\delta=0.1$ (Failure Case)}

\blockB{To determine how much Janet makes every day at the farmers' market, we need to follow these steps:}

\blockB{1. Calculate the total number of eggs laid by Janet's ducks per day.}

\blockB{2. Subtract the number of} \blockA{eggs Janet eats and bakes for her friends from the total number of eggs laid.}

\blockA{3. Multiply the number of eggs sold by the price per egg to find her daily earnings.}

\blockA{First, let's calculate} \blockB{the total number of eggs laid by Janet's ducks per day:}
\[
\blockB{16 \text{ eggs}}
\]

\blockB{Next, we subtract the number of eggs Janet eats and bakes for her friends:}
\[
\blockB{16 \text{ eggs} - 3 \text{ eggs} - 4 \text{ eggs} = 11 \text{ eggs}}
\]

\blockB{Now, we multiply the number of} \blockA{eggs sold by the price per egg to find her daily earnings:}
\[
\blockA{11 \text{ eggs} \times 2 \text{ dollars/egg} = 22 \text{ dollars}}
\]

\blockA{Therefore, the amount} \blockB{Janet makes every day at the farmers' market is:}
\[
\boxed{\blockB{22}}
\]

\end{graybox}

\begin{graybox}[title={Case 2: LLaDA-8B, GSM8K, Generation Length 256}]

\textbf{Question}

A robe takes 2 bolts of blue fiber and half that much white fiber.  How many bolts in total does it take?

\medskip

\textbf{Ground-truth Answer}

It takes 2/2=<<2/2=1>>1 bolt of white fiber\\
So the total amount of fabric is 2+1=<<2+1=3>>3 bolts of fabric

\noindent\rule{\linewidth}{0.5pt}

\textbf{GeoBlock, max Block Length = 32, $\delta=0.1$}

\blockB{To determine the total number of bolts required for the robe, we need to add the amount of blue fiber and} \blockA{the amount of white fiber.}

\blockA{Step 1: Identify the amount of} \blockB{blue fiber.}

\blockB{The robe takes 2 bolts of blue fiber.}

\blockB{Step 2: Determine the amount} \blockA{of white fiber.}

\blockA{The robe takes half as much white fiber as blue fiber. Therefore, we calculate} \blockB{the amount of white fiber as follows:}
\[
\blockB{\text{Amount of white fiber} = } \blockB{\frac{1}{2} \times 2 } \blockA{= 1 \text{ bolt}}
\]

\blockB{Step 3: Calculate the total number of bolts by adding the amount of blue fiber and the amount of} \blockA{white fiber.}
\begin{align*}
&\blockA{\text{Total number of bolts}} \\
&= \blockA{\text{Amount of blue fiber} }+  \blockB{\text{Amount of white fiber}}
\end{align*}
\[
\blockB{\text{Total number of bolts} = 2 + 1 = } \blockA{3}
\]

\blockA{Therefore, the total number of bolts required for the robe is } \boxed{\blockB{3}}\blockB{.}\\

\textbf{Block length}: $19.7 \pm 4.4$. \textbf{Number of blocks}: 13.

\noindent\rule{\linewidth}{0.5pt}

\textbf{GeoBlock, max Block Length = 64, $\delta=0.05$}

\blockB{To determine the total number of bolts required for the robe, we need to add the amount of blue fiber to the amount of white fiber.}

\blockA{Step 1: Identify the amount of blue fiber.}

\blockA{The robe takes 2 bolts of blue fiber.}

\blockA{Step 2: Determine the amount of} \blockB{white fiber.}

\blockB{The robe takes half as much white fiber as blue fiber. Since it takes 2 bolts of blue fiber, the amount of white fiber is:}\\
\[
\blockB{\frac{1}{2} \times 2 = 1 \text{ bolt}}
\]

\blockA{Step 3: Calculate the total number of} \blockB{bolts.}

\blockB{Add the amount of blue fiber to the amount of white fiber:}
\[
\blockB{2 + 1 = 3}
\]
\blockB{Therefore, the} \blockA{total number of bolts required is } \boxed{\blockA{3}}\blockA{.}\\

\textbf{Block length}: $36.4 \pm 10.6$. \textbf{Number of blocks}: 7.

\end{graybox}

\end{document}